\documentclass[sigconf]{acmart}
\AtBeginDocument{%
  }

\setcopyright{acmlicensed}
\copyrightyear{2026}
\acmYear{2026}
\acmDOI{XXXXXXX.XXXXXXX}
\acmConference[DAI 2026]{DAI}{November 29--December 2, 2026}
{Hong Kong}
\acmISBN{978-1-4503-XXXX-X/2026/06}

\usepackage{algorithm}
\usepackage{algorithmic}
\usepackage{dblfloatfix}
\usepackage{cuted}
\usepackage{caption} 
\usepackage{multirow}

\begin{document}

\title{Collab-Solver: Collaborative Solving Policy Learning for Mixed-Integer Linear
Programming}

\author{Siyuan Li}
\affiliation{%
  \institution{Harbin Institute of Technology}
  \city{Harbin}
  \country{China}
}
\email{siyuanli@hit.edu.cn}

\author{Yifan Yu}
\affiliation{%
    \institution{Harbin Institute of Technology}
  \city{Harbin}
  \country{China}
  }

\author{Zhihao Zhang}
\affiliation{%
  \institution{Harbin Institute of Technology}
  \city{Harbin}
  \country{China}
}

\author{Mengjing Chen}
\affiliation{%
 \institution{Huawei Noah's Ark Lab}
 \city{Beijing}
 \country{China}
 }

\author{Fangzhou Zhu}
\affiliation{%
  \institution{Huawei Noah's Ark Lab}
 \city{Beijing}
 \country{China}
  }

\author{Tao Zhong}
\affiliation{%
  \institution{Huawei Noah's Ark Lab}
 \city{Beijing}
 \country{China}
  }

\author{Jianye Hao}
\affiliation{%
  \institution{Huawei Noah's Ark Lab}
 \city{Beijing}
 \country{China}
  }

\author{Peng Liu}
\affiliation{%
  \institution{Harbin Institute of Technology}
  \city{Harbin}
  \country{China}
  }
\email{pengliu@hit.edu.cn}

\renewcommand{\shortauthors}{Trovato et al.}

\begin{abstract}
Mixed-integer linear programming (MILP) is a fundamental problem in combinatorial optimization, and conventional MILP solving primarily relies on carefully designed heuristics within the branch-and-bound framework.
Driven by the strong capabilities of neural networks, recent research is exploring machine learning techniques alongside conventional MILP solving. 
    Although learning-based MILP methods have shown great promise, existing work typically learns policies for individual modules in MILP solvers in isolation, without considering their interdependency, which limits both solving efficiency and solution quality.
    To address this limitation, we propose Collab-Solver, a novel multi-agent-based policy learning framework for MILP that enables collaborative policy optimization across multiple modules.
    Specifically, we formulate the collaboration between cut selection and branching in MILP solving as a Stackelberg game.
    Under this formulation, we develop a two-phase learning paradigm to stabilize collaborative policy learning: the first phase performs data-communicated policy pretraining, and the second phase further orchestrates the policy learning for various modules. 
    Extensive experiments on both synthetic and large-scale real-world MILP datasets demonstrate that the jointly learned policies significantly improve solving performance. Moreover, the policies learned by Collab-Solver have also demonstrated excellent generalization abilities across different instance sets.
\end{abstract}

%
%





\maketitle

\section{Introduction}
\begin{figure}[htbp]
    \centering
    \includegraphics[width=0.75\linewidth]{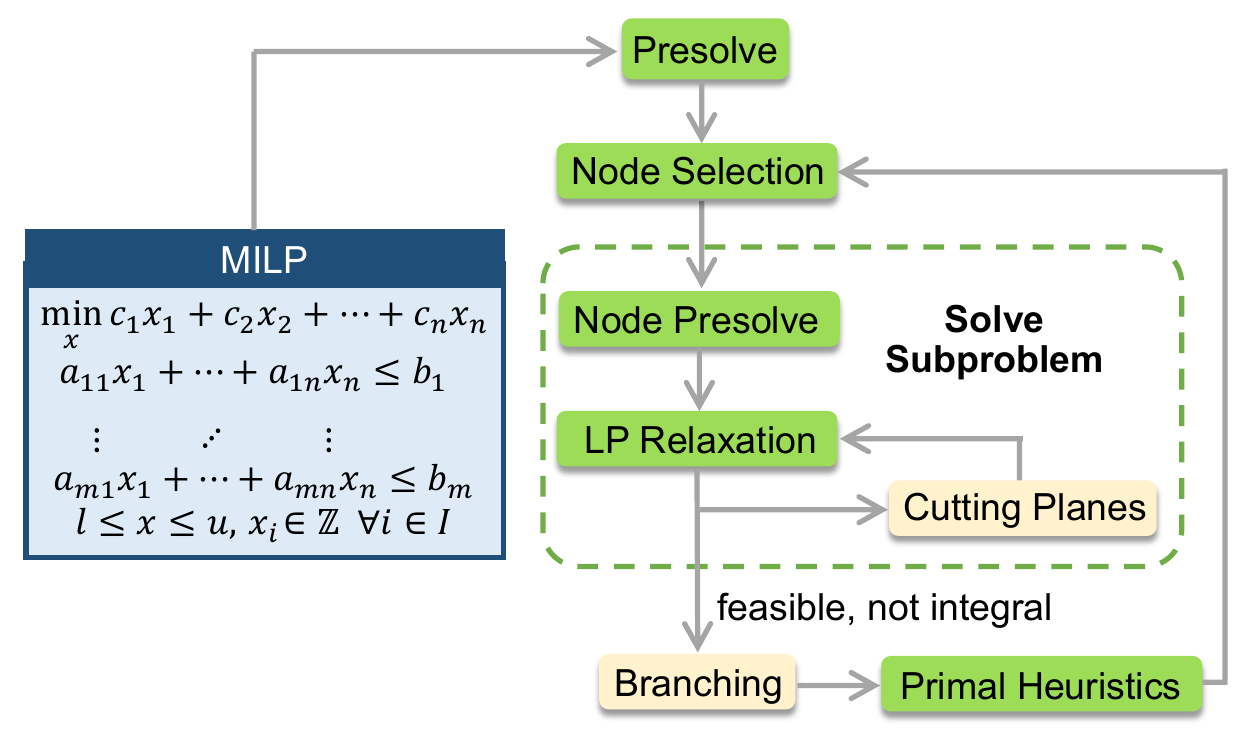}
    \caption{Flowchart of the MILP solving loop, which involves multiple closely related modules.} 
    \vspace{-10pt}
    \label{fig1}
\end{figure}

Mixed-integer linear programming (MILP) refers to linear objective optimization problems with linear constraints and integer constraints, i.e., some or all of the variables are integers \cite{bixby2004mixed}. 
Despite its broad applicability, MILP is extremely challenging to solve, as most instances are NP-hard.
Modern solvers such as SCIP \cite{bestuzheva2021scip}, CPLEX \cite{bliek1u2014solving}, and Gurobi \cite{pedroso2011optimization} are built upon the branch-and-bound (B\&B) framework \cite{land2010automatic}, which integrates multiple tightly coupled algorithmic modules to efficiently explore the search space.
As illustrated in Figure~\ref{fig1}, such MILP solvers integrate multiple tightly coupled algorithmic modules, including presolve, cut generation and selection, node selection, and branching \cite{achterberg2009scip}, to efficiently explore the B\&B search space.
Traditionally, solver performance has relied on effective heuristics for individual modules \cite{anand2017comparative,berthold2018parallelization}.
However, designing these heuristics requires substantial manual effort and often lacks robustness when transferred to new problem domains.


Recent advances in deep learning have enabled learning-based MILP solving, where neural network-based policies are trained to replace or enhance solver heuristics.
Prior work has demonstrated promising results in learning branching \cite{scavuzzo2022learning,shanmugasundaram2022intelligent}, node selection \cite{zhang2023reinforcement}, and cut selection \cite{paulus2022learning,10607926}, achieving improved performance over handcrafted rules \cite{zhang2023survey}.
Existing studies have explored learning individual solver modules and organizing multiple modules within multi-agent frameworks. However, combining learned modules does not by itself resolve their sequential dependence. In branch-and-cut, cut selection modifies the LP relaxation and therefore changes the observations and candidate variables encountered by the downstream branching policy. Consequently, independently trained policies may suffer from a deployment mismatch, while concurrent updates introduce non-stationarity because each policy changes the learning environment of the other. These challenges motivate a collaboration mechanism that explicitly coordinates data generation and policy updates across solver modules.

To address this problem, we propose Collab-Solver, a multi-agent collaborative policy learning framework for MILP solving.
In this work, we focus on the collaboration between the cut selection module and the branching module.
As shown in Figure \ref{fig1}, these two modules both paly a key role in MILP solving, and the cutting planes substantially influence the node to be branched.
We formulate their sequential decision-making process as a \textit{Stackelberg game} \cite{1994A} and develop a two-phase learning paradigm for stable joint optimization.
Specifically, we introduce a data communication mechanism to facilitate information sharing between modules, and then apply a two-timescale update rule to mitigate non-stationarity during concurrent policy learning.
The proposed Collab-Solver has been extensively evaluated on eight NP-hard benchmark datasets. Extensive experiment results demonstrate that Collab-Solver significantly outperforms existing learning-based MILP methods. 
Next,  we demonstrate the superior generalization ability of our method on out-of-distribution test datasets.
Besides, we conduct various ablation studies to provide further insights into Collab-Solver and show its ability to extend to more module collaboration. 
The contributions of this work can be summarized as below.
\begin{itemize}
    \item We formulate the sequential interaction among solver modules as a hierarchical leader-follower learning problem, where upstream policies alter the state distributions observed by downstream policies.
    \item We develop a two-phase collaboration scheme that first constructs policy-conditioned training data and then stabilizes joint policy optimization through a two-timescale update rule.
    \item We validate the framework on cut selection and branching across six MILP benchmarks, and further extend it to a three-module hierarchy involving predict-and-search, cut selection, and branching.
\end{itemize}

\section{Background}
In this section, we first introduce the preliminaries of MILP, and then introduce the branch-and-cut algorithm. 
\subsection{Mixed Integer Linear Programming}
An MILP instance is defined as
\begin{equation}
z^*=\min_{x\in\mathbb{R}^n}\{c^Tx \mid Ax\le b,\ x_i\in\mathbb{Z},\ \forall i\in\mathcal I\},
\label{eq1}
\end{equation}
where $\mathcal I$ denotes the set of variables constrained to take integer values.
MILP solvers typically rely on multiple interacting components, including cutting planes and branching, which are tightly coupled in the solving process.

\subsection{Branch-and-Cut}



Given Problem \eqref{eq1}, we can drop all the integer constraints to obtain its LP relaxation,
\begin{equation}
z_{LP}^*\triangleq\min_{x\in\mathbb{R}^n}\{c^Tx|Ax\leq b \}.
\label{eq2}
\end{equation}
Since Problem \eqref{eq2} expands the feasible set of Problem \eqref{eq1}, it holds that $z_{LP}^*\leq z^{*}$. The value $z_{LP}^{*}$ provides a valid lower bound on the optimal objective value of the original MILP and is referred to as a {dual bound}.  
With the LP relaxation in \eqref{eq2}, cutting planes (cuts) are linear inequalities added to the LP relaxations to tighten them without removing any integer feasible solutions of Problem \eqref{eq1}. By strengthening the LP relaxation, cutting planes improve the quality of the LP objective value and lead to tighter dual bounds. Cuts generated by the solver are added in consecutive rounds. Each round includes three steps: (i) solving the current LP relaxation; (ii) generating a pool of candidate cuts; (iii) selecting a subset and adding it to the current LP relaxation to obtain the next one. \textbf{Cut selection} (i.e., the third step in cutting planes) is quite important, as it directly influences the constraints and determines how effectively the dual bound is tightened.
Following previous work \cite{wanglearning,10607926}, we primarily investigate cutting-plane generation and selection for a single round at each node, where these nodes are arbitrary nodes specified by the base solver.
In modern MILP solvers, cutting planes are typically combined with the B\&B algorithm, known as branch-and-cut (B\&C). The B\&B algorithm performs implicit enumeration by building a search tree, where each node represents a subproblem of the original problem in \eqref{eq1}. The B\&B algorithm selects a leaf node of the tree and solves its LP relaxation, yielding a node-specific dual bound. Let $x^*$ be the optimal solution of the LP relaxation. If $x^*$ violates the integrality constraints in Problem \eqref{eq1}, the leaf node is augmented with the constraints
\begin{equation}
x_i \leq \lfloor x_i^* \rfloor \text{\ or\ } x_i \geq \lceil x_i^* \rceil,
\end{equation}
decomposing it into two subproblems.
Choosing the variable $x_i$ at every B\&B iteration is called \textbf{branching}.
If $x^*$ is a solution of Problem \eqref{eq1}, we obtain an upper bound on $z^*$, called a primal bound. 
The gap between the global primal bound and the global dual bound is called the primal-dual gap (PD gap). The solving terminates when the PD gap is $0$ or the solving time limit is reached. 
In the B\&C algorithm, adding cutting planes is alternated with branching, and they are strongly correlated.

\begin{figure*}[t!]
    \centering
    \includegraphics[width=0.6\linewidth]{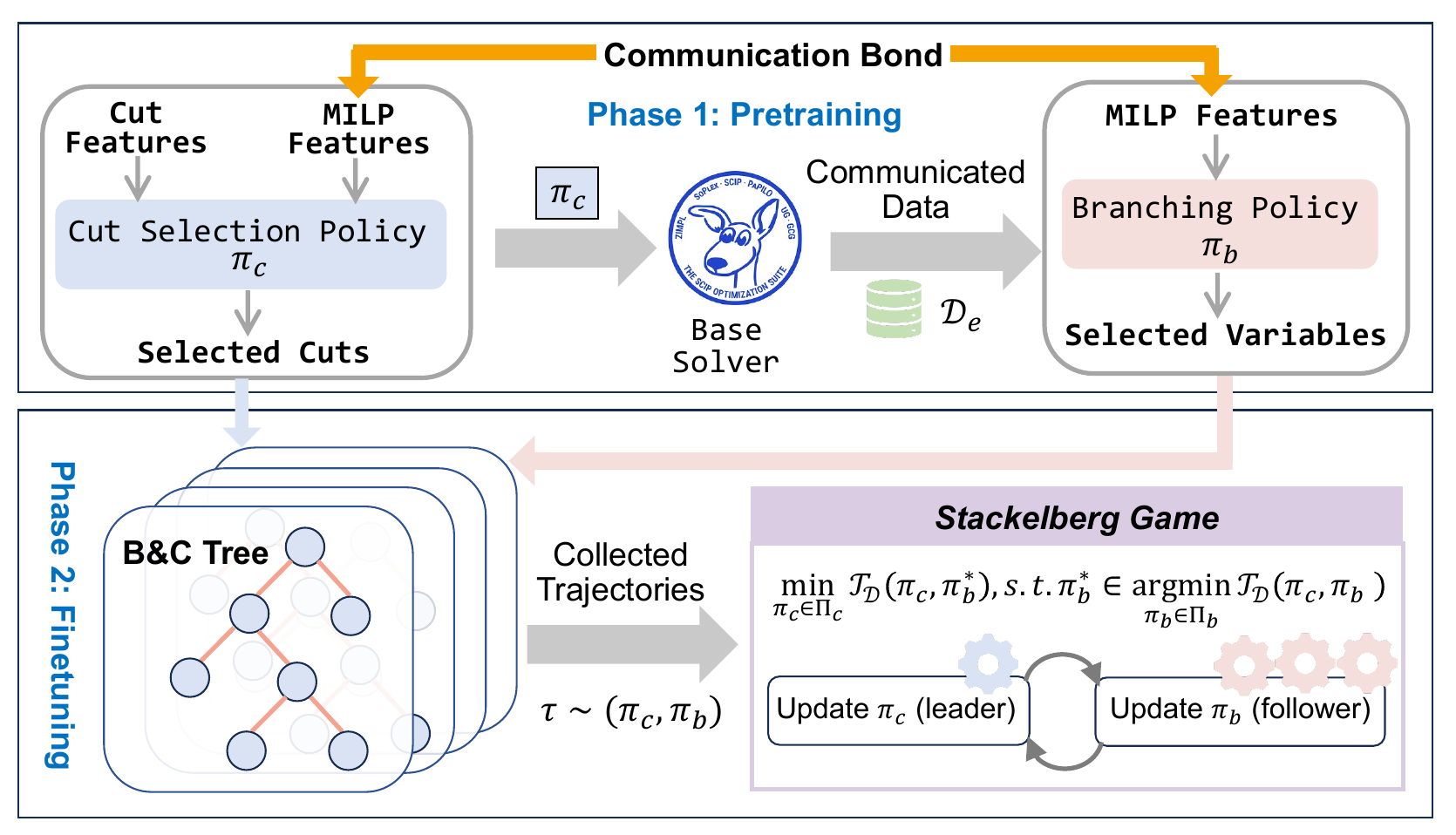}
    \caption{The Collab-Solver framework. The upper part illustrates the first learning phase: data communicated pretraining, and the lower part describes the second learning phase: concurrent joint finetuning.}
    \label{fig2}
\end{figure*}

\section{Related Work}
Mixed-integer linear programming (MILP) is a fundamental and challenging problem in combinatorial optimization \cite{floudas2005mixed}, with a wide range of real-world applications, including supply chain management \cite{paschos2014applications}, production planning \cite{junger200950}, and scheduling \cite{chen2010integrated}.
Due to its inherent NP-hardness, early research on MILP solving primarily focused on the design of effective heuristics based on extensive human expertise \cite{achterberg2007constraint}.
Modern MILP solvers, such as SCIP \cite{bestuzheva2021scip}, CPLEX \cite{bliek1u2014solving}, and Gurobi \cite{pedroso2011optimization}, are primarily built on the B\&B framework, which integrates multiple algorithmic components to explore the combinatorial search space efficiently.
A notable characteristic of modern solvers is the presence of many hyperparameters, whose configurations can significantly affect solving performance. To reduce the reliance on manual tuning, a variety of automatic hyperparameter optimization methods have been proposed. Among them, SMAC3 \cite{lindauer2022smac3} is a representative tool based on Bayesian optimization that has been widely adopted for solver configuration.

With the rapid development of deep learning, recent studies have explored learning-based approaches to replace manually designed heuristics and further alleviate the hyperparameter tuning burden \cite{zhang2023survey}.
Learning-based policies have been successfully applied to different solver modules, demonstrating improved performance and reduced dependence on expert knowledge. Representative applications include node selection \cite{zhang2023reinforcement,zhang2025learning}, primal heuristics \cite{paulus2024learning}, large-neighborhood search \cite{yuan2025btbs}, and neural model reduction or solution generation \cite{liu2025apollo,li2025constraint,li2025fmip}.
Among existing learning-based MILP approaches, methods for cut selection and branching \cite{scavuzzo2024machine} are most closely related to our work.
For cut selection, previous studies have proposed reinforcement learning--based approaches \cite{wanglearning,wang2024enhancing}, imitation learning techniques \cite{paulus2022learning}, and hybrid imitation--reinforcement learning \cite{m2026miracle}; recent work also studies dynamic separator configuration and the sample complexity of learning cuts \cite{ye2026dynamic,khalife2025hard}.
In particular, HEM \cite{10607926} introduces a hierarchical model in which a high-level policy controls the proportion of cuts to be selected, while a low-level policy determines the specific cuts to be added.
Regarding branching, which can also be naturally formulated as a sequential decision-making problem, similar learning paradigms have been adopted \cite{shanmugasundaram2022intelligent,scavuzzo2022learning,qu2022yordle,strang2026markov}.
A representative work in this direction is learning to branch \cite{gasse2019exact}, which leverages graph convolutional neural networks (GCNNs) to parameterize branching decisions and has become a standard backbone for many subsequent learning-based MILP methods; recent work further explores attention-based backbones for more expressive MILP representations \cite{huang2026general}.
More recently, several works have focused on improving the generalization ability of learned MILP policies through multi-task and cross-domain learning, enabling transfer to previously unseen instances \cite{cai2025multi,huang2024distributional,pu2026rome}, with complementary sample-complexity guarantees established for score-based B\&C policies \cite{cheng2026generalization}.
As generalization with multi-task learning is not the primary focus of this work, we do not elaborate further on this line of research.

Despite the impressive performance achieved by existing learning-based MILP approaches, most methods train policies for different solver modules independently, without explicitly modeling their interdependencies; moreover, locally accurate cut or branching scores need not yield efficient global B\&C trees \cite{cheng2026theoretical}.
In practice, however, solver modules such as cut selection and branching are tightly coupled, and decisions made by one module can significantly influence the effectiveness of others.
As a result, independently learned policies often fail to collaborate effectively, leading to suboptimal overall solving efficiency.
A recent work \cite{jing2024multi} also explores collaborative policy learning for MILP.
Compared to this approach, Collab-Solver features a more tightly coupled collaboration mechanism by concurrently learning multiple policies with explicit data communication and joint fine-tuning.
Moreover, our method removes the restriction that cuts are confined to the root node, allowing cut selection and branching to be performed sequentially throughout the search process.
In contrast, the method in \cite{jing2024multi} applies cuts only at the root node.
In the ablation studies of the experimental section, we further compare Collab-Solver with a variant that removes fine-tuning and data communication, which is approximately consistent with the method proposed in \cite{jing2024multi}.


\section{Methodology}
Although extensive literature has investigated learning-based methods to replace or enhance handcrafted heuristics in MILP solving, it primarily focuses on individual components and neglects the synergy among modules, leading to poor solving performance. 
To address this problem, we propose a novel multi-agent-based solving policy learning paradigm called Collab-Solver, enabling effective collaboration between closely related modules.
Particularly, Collab-Solver focuses on the collaboration between the cut selection policy and the branching policy. 
It is noteworthy that the proposed methodology can be extended to collaborations among other modules in MILP solvers, as briefly discussed at the end of Section \ref{sec31} and validated in the experiment section. Next, we first present the problem formulation and then introduce the two learning phases in the Collab-Solver scheme. 


\subsection{Problem Formulation}
\label{sec31}
 The cutting plane and branching modules can be formulated as two agents, denoted the cutting agent and the branching agent respectively. 
As shown in Figure \ref{fig1}, these two agents sequentially make decisions with a shared objective, i.e., minimizing the solving time and improving the solution quality. We propose to formulate the collaborative policy learning problem with a game-theoretic model, Stackelberg game \cite{1994A}. 
Specifically, the solving process is formulated as a finite-horizon partially-observable Markov game $\mathcal{G}=(S,A_c, A_b,P,O_c, O_b,r)$, where $S$ denotes the state space for MILP solving, $A_c$ and $A_b$ denote the action spaces for the cutting agent and the branching agent respectively, and $r$ is the reward function related to the optimization objective.
The cutting agent observes cut-related features $o_c \in O_c$, while the branching agent observes MILP instance features $o_b \in O_b$ encoded as a bipartite graph.
At each decision step, the cutting agent selects a subset of generated cuts $a_c \in A_c$ to update the current LP relaxation based on $(o_c, o_b)$.
If the resulting LP solution violates integrality constraints, the branching agent selects a variable for branching based on $o_b$, generating subproblems.
The solver state then transitions according to $P(s' \mid s, a_c, a_b)$, and both agents receive new observations and rewards.
Since branching decisions depend on the outcome of cut selection, we treat the cutting agent as the leader and the branching agent as the follower.
The collaborative learning objective is formulated as the following bi-level optimization problem:
\begin{equation}
    \min_{\pi_c\in \Pi_c} \  \mathcal{T}_\mathcal D(\pi_c, \pi_b^*)  \text{s.t. } \  \pi_b^* \in \arg \min_{\pi_b\in \Pi_b} \mathcal{T}_\mathcal D(\pi_c, \pi_b),
\label{eq_game}
\end{equation}
where $\mathcal{T}_{\mathcal D}(\pi_c, \pi_b)
= \mathbb{E}_{\mathcal D}\!\left[\sum_t r(s^t, a_c^t, a_b^t)\right]$
denotes the expected cumulative reward over the training dataset $\mathcal D$.
In the standard setting, the reward is defined as the negative solving time.
For extremely challenging instances where solving time becomes less informative, we instead use the PD gap as the reward signal to better reflect solution quality.
Based on this formulation, we develop a practical two-phase learning paradigm for stable collaboration between the cutting and branching policies.
Moreover, the proposed bi-level optimization framework can be naturally extended to multi-module collaboration under a hierarchical Stackelberg game formulation.

\subsection{Data Communicated Pretraining}


To reduce the deployment mismatch between sequential modules, we pretrain the branching policy $\pi_b$ on states induced by the learned cut-selection policy $\pi_c$. This data-communicated pretraining ensures that the follower observes LP relaxations and branching candidates consistent with those encountered during deployment. We next describe the pretraining of the cut-selection and branching policies; the pseudocode algorithm is provided in Appendix C.1.

\subsubsection{Cut Selection Policy Pretraining}

\begin{figure}[h]
    \centering
    \includegraphics[width=0.95\linewidth]{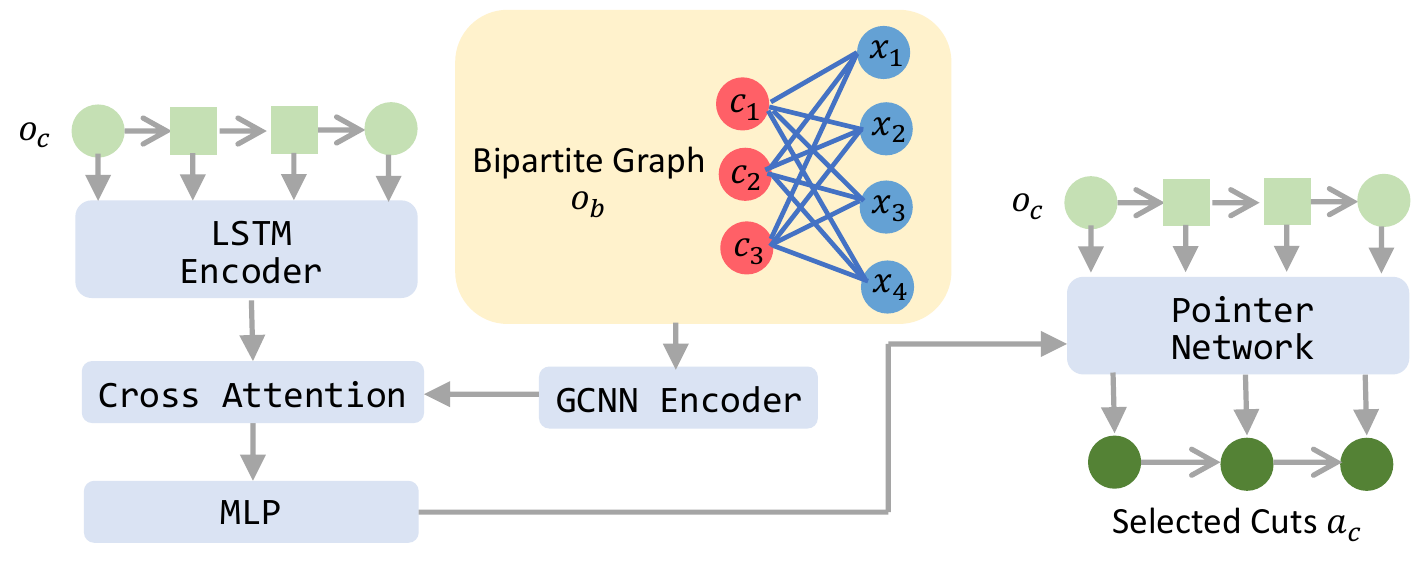}
    \caption{Network structure of $\pi_c$, which is composed of LSTM encoder, GCNN encoder, MLP, and pointer network.}
    \label{fig_pi_c}
\end{figure}

Inspired by the previous cut selection approach HEM \cite{wanglearning}, we propose to first learn the cutting policy $\pi_c$ with the policy gradient algorithm in reinforcement learning \cite{sutton2018reinforcement}. 
Different from HEM, since we consider the collaboration between the cutting policy $\pi_c$ and the branching policy $\pi_b$, the input for $\pi_c$ not only involves the candidate cut features $o_c$, but also includes the MILP features $o_b$, as shown in the upper left of Figure \ref{fig2}. 
Note that the MILP features $o_b$ serve as the communication bond between the two policies, and thus the cut selection policy takes the form of $\pi_c(a_c|o_c\circ o_b)$. The  $\circ$ notation represents that $o_c$ and $o_b$ are both fed into the network $\theta$, but with different entries, as shown in Figure \ref{fig_pi_c}, which is beyond a simple concatenation. 
Since there are multiple candidate cuts, $o_c$ is a sequence of input data. The MILP features $o_b$ that describe the current node on the B\&C tree are typically represented as a bipartite graph.

The network structure of $\pi_c$ is illustrated in Fig. \ref{fig_pi_c}. Cut feature $o_c$ is first processed by an LSTM encoder to capture the sequential dependencies and contextual relationships among candidate cuts. Meanwhile, the bipartite MILP features $o_b$, consisting of variable and constraint nodes, are represented as a bipartite graph and encoded using a GCNN \cite{zhang2019graph}, which extracts structural information from the current LP relaxation. $o_b$ denotes the MILP features, reflecting the interactions between variables and constraints, whereas the cut features $o_c$ describe local properties of individual candidate cuts to be selected. To effectively integrate these heterogeneous sources of information, we employ a cross-attention network \cite{hou2019cross}, which enables the model to dynamically attend to relevant global solving states when evaluating each candidate cut. The fused representations are then passed through a multi-layer perceptron (MLP) to produce a joint embedding that summarizes both the global optimization context and the cut-specific information. 
Finally, since the number of candidate cuts varies across nodes in the branch-and-bound tree, we adopt a pointer network  \cite{vinyals2015order} to deal with action spaces of varying sizes. 
The pointer network outputs a probability distribution over the candidate cuts, from which the action $a_c$ is sampled. 
Details of the observations and actions for the cutting and branching agents are provided in Appendix~\ref{app1}. The parameters $\theta$ of $\pi_c$ are optimized by maximizing the expected cumulative reward over trajectories using policy gradient as follows.
\begin{equation}
\nabla_\theta J_c(\theta) = \mathbb{E}_{s\sim\mu, a_c\sim\pi_c(\cdot|o_c\circ o_b)}[\nabla_\theta \log \pi_c(a_c|o_c\circ o_b)r(s,a_c)],
\label{eq4}
\end{equation}
where $\mu$ is the initial state distribution. $r(s,a_c)$ represents the reward of selecting the cuts $a_c$ at the state $s$, which serves as a simplification for $r(s, a_c, a_b)$, as in this stage the branching policy $\pi_b$ is not learned, but set as default in the SCIP solver \cite{bestuzheva2021scip}.
For relatively easy MILP instances that can be solved within a short time limit, we set the reward $r$ to the negative of the solving time. Non-zero rewards are only given when the solving process ends, and the other rewards are zero. For difficult instances, the reward is based on PD gaps, where difficulty is assessed by computing the average solving time over a small subset of the training data.

\subsubsection{Branching Policy Pretraining}

After learning $\pi_c$, we replace the cut selection heuristic in the SCIP\footnote{We utilize SCIP as the base solver since it is open-sourced and flexible to modify. In contrast, the commercial solvers, such as Gurobi and CPLEX, do not have open APIs for cut selection and branching \cite{lawless2021reinforcement}.} solver with $\pi_c$, and utilize the \textit{strong branching} rule as the branching policy in the B\&C algorithm to generate an expert dataset $\mathcal{D}_e=\{(o_b^i,a_b^i)\}_{i=1}^N$ for branching, where $o_b^i$ denotes the observation for the branching agent, and $a_b^i$ represents the variables selected by the strong branching rule. 
The data communication between these two agents is conducted in an implicit manner, where the cuts $a_c$ selected by $\pi_c$ are added as constraints to the current MILP, so the actions taken by $\pi_c$ are implicitly included in the bipartite graph $o_b$.  
Therefore, the expert dataset $\mathcal{D}_e$ for branching is inherently conditioned on the learned cut selection policy $\pi_c$. 
Since the strong branching algorithm usually produces the smallest tree but suffers a high computation cost, the expert dataset $\mathcal{D}_e$ is employed to optimize $\pi_b$ with the behavioral cloning loss $\mathcal{L}(\psi)$ \cite{pomerleau1991efficient}, and thus $\pi_b$ can achieve effective branching with neural network inference speed,
\begin{equation}
\mathcal{L}(\psi) = -\frac{1}{N}\sum_{(o_b^i,a_b^i)\in\mathcal{D}_e}\log \pi_{b}(a_b^i|o_b^i).
\label{imi_loss}
\end{equation}
$\psi$ denotes the parameter of the branching policy $\pi_b(a_b|o_b)$, which is formulated with the observation $o_b$ to output the selected variable $a_b$. 
Observation $o_b^i$ is formulated as a bipartite graph with one part as constraints and the other as variables, the same as the MILP features of the inputs of $\pi_c$.
To tackle the graph features $o_b$, the network structure of $\pi_b$ is designed as a GCNN followed by MLPs, similar to the previous work \cite{gasse2019exact}.
The hyperparameters in the training process are listed in  Appendix~\ref{app2}, and the pseudocode of data communicated pretraining is presented in Appendix~\ref{app:Pseudocode-1}.


\subsection{Concurrent Joint Finetuning}

With pretrained $\pi_b$ and $\pi_c$, Collab-Solver further jointly finetunes these two policies in an online manner to achieve effective collaboration.
Specifically, the two pretrained policies are employed to solve one instance, and the generated training datasets are used to concurrently optimize these two policies.
However, the concurrent learning of two related policies inherently induces the non-stationary issue. Therefore, we develop a two-timescale update rule, where the two policies have different update frequencies. 
\begin{itemize}
    \item \textbf{Slow timescale.}  Updating the leader policy $\pi_c$ once every $\omega_c$ instances. The leader $\pi_c$ can be regarded as a generator, which generates the node to be branched. As the leader's decision has a huge influence on this game, the leader's policy $\pi_c$ needs to be updated slowly.
    \item \textbf{Fast timescale.} Updating the follower policy $\pi_b$ once every $\omega_b$ instances, where $\omega_b<\omega_c$. The follower's decision is dependent on the leader's decision, and thus $\pi_b$ needs to adapt fast to the changes of $\pi_c$.
\end{itemize}
With this two-timescale update mechanism, the online finetuning process is enhanced with smoothness, i.e., the optimization of one policy is less likely to hurt the other, so that the collaborative update is stabilized. Without such a two-timescale update mechanism, simultaneous updates of the two policies may lead to instability, or even degraded performance, as changes in one policy may negatively interfere with the learning of the other.

\begin{table*}[t!]
    \centering
    \begin{tabular}{ccccccc}
        \toprule\toprule
        &\multicolumn{2}{c}{Set Covering}  & \multicolumn{2}{c}{Max Independent Set} & \multicolumn{2}{c}{Combinatorial Auction} \\
       & \multicolumn{2}{c}{(n = 1000, m = 500)}  & \multicolumn{2}{c}{(n = 500, m = 1953)} & \multicolumn{2}{c}{ (n = 500, m = 192)} \\
        \midrule
        Method & Time(s) $\downarrow$ & PD Integral $\downarrow$  & Time(s) $\downarrow$ & PD Integral $\downarrow$  & Time(s) $\downarrow$ & PD Integral $\downarrow$ \\
         \cmidrule(lr){2-3} \cmidrule(lr){4-5} \cmidrule(lr){6-7}
        SCIP     & 4.24 $\pm$ 0.08          & 53.58 $\pm$ 1.06       & 
         4.27 $\pm$ 0.19          & 33.63 $\pm$ 1.68       & 
        1.65 $\pm$ 0.07          & 14.03 $\pm$ 0.71\\
        SMAC3     & 6.40 $\pm$ 0.06          & 58.97 $\pm$ 1.29
        & 
         2.66 $\pm$ 0.10          & 21.33 $\pm$ 1.99       & 
        1.66 $\pm$ 0.11         & 13.09 $\pm$ 0.65\\
        HEM     & 2.74 $\pm$ 0.06          & 52.66 $\pm$ 1.20      & 
        2.40 $\pm$ 0.12          & 22.60 $\pm$ 1.30      & 
        1.58 $\pm$ 0.08          & 13.82 $\pm$ 0.61  \\ 
        GCNN-B  & 4.03 $\pm$ 0.29          & 49.76 $\pm$ 2.19      & 
         4.98 $\pm$ 1.07          & 28.27 $\pm$ 5.05      & 
         0.92 $\pm$ 0.01          & 11.77 $\pm$ 0.19     \\
         RL-B     &  6.68 $\pm$ 5.92          & 81.33 $\pm$ 14.11       & 
         6.99 $\pm$ 4.57          & 41.64 $\pm$ 6.55       & 
        1.49 $\pm$ 0.84         & 12.29 $\pm$ 2.63\\
        Ours    & $\textbf{2.45 $\pm$ 0.05}$          & $\textbf{49.72 $\pm$ 0.87}$      & 
        $\textbf{1.41 $\pm$ 0.49}$          & $\textbf{13.19 $\pm$ 0.34}$      & 
         $\textbf{0.78 $\pm$ 0.03}$         & $\textbf{10.07 $\pm$ 0.46}$ \\
        \bottomrule
        \\
        \toprule\toprule
        &\multicolumn{2}{c}{Capacitated Facility Location}  & \multicolumn{2}{c}{Mixed Integer Knapsack} & \multicolumn{2}{c}{Production Planning} \\
        & \multicolumn{2}{c}{ (n = 10100, m = 10203)}  & \multicolumn{2}{c}{(n = 413, m = 346)} & \multicolumn{2}{c}{ (n = 3582.25, m = 5040.42)} \\
        \midrule
        Method & Time(s) $\downarrow$ & PD Integral $\downarrow$  & Time(s) $\downarrow$ & PD Integral $\downarrow$  & Time(s) $\downarrow$ & PD Integral $\downarrow$ \\
        \cmidrule(lr){2-3} \cmidrule(lr){4-5} \cmidrule(lr){6-7}
        SCIP     & 80.23 $\pm$ 3.19          & 300.28 $\pm$ 20.18      & 
         70.42 $\pm$ 4.23          & 150.78 $\pm$ 17.18      & 
         198.63 $\pm$ 1.72         & 11664.33 $\pm$ 17.67\\
         SMAC3     & 68.91 $\pm$ 4.10          & 297.10 $\pm$ 23.31       & 
         64.46 $\pm$ 6.22          & 226.76 $\pm$ 26.23       & 
        160.60 $\pm$ 2.11         & 11599.73 $\pm$ 19.65\\
        HEM      & 80.97 $\pm$ 7.27          & 305.32 $\pm$ 44.64      & 
         68.73 $\pm$ 6.49          & 195.18 $\pm$ 22.75      & 
         \textbf{138.08 $\pm$ 0.13}         & 8648.01 $\pm$ 18.92 \\
        GCNN-B   & 69.87 $\pm$ 3.74          & 297.42 $\pm$ 24.89      & 
         82.18 $\pm$ 3.09          & 172.48 $\pm$ 7.85       & 
        153.64 $\pm$ 1.15         & 8108.12 $\pm$ 89.74     \\
        RL-B     &  78.14 $\pm$ 10.09          & 290.62 $\pm$ 35.68       & 
         85.45 $\pm$ 7.99          & 237.74 $\pm$ 23.47       & 
        198.38 $\pm$ 2.17         & 11233.49 $\pm$ 118.87\\
        Ours     & $\textbf{57.10 $\pm$ 5.87}$          & $\textbf{282.99 $\pm$ 5.54}$      & 
         $\textbf{46.17 $\pm$ 7.75}$          & $\textbf{117.78 $\pm$ 13.52}$      & 
         \textbf{138.85 $\pm$ 0.80}         & $\textbf{6812.83 $\pm$ 78.78}$\\
        \bottomrule
    \end{tabular}
    \caption{Comparative experiment results. The best performance is marked in bold. Each experiment has been run with $5$ random seeds, and the mean and standard deviation are listed above. For Collab-Solver, $3$ out of $10$ test instances in the real-world dataset (Production Planning) have reached the time limit, whereas in the other datasets, the time limits have not been hit.
    }
    \label{tab1}
\end{table*}
As the models are continuously updated in this phase, offline imitation learning of $\pi_b$ is no longer possible.
Instead, we propose to adopt the policy gradient method to fine-tune $\pi_b$ as follows.
\begin{equation}
\nabla_\psi J_b(\psi) = \mathbb{E}_{\tau\sim (\pi_c, \pi_b)}[\nabla_\psi \log \pi_{b}(a_b|o_b)r(s,a_c, a_b)],
\label{eq6}
\end{equation}
where $\tau$ denotes the trajectory generated by solving one instance with $\pi_b$ and $\pi_c$ under the $\epsilon$-greedy exploration strategy \cite{sutton2018reinforcement}. The reward $r$ in Equation \eqref{eq6} is consistent with that in Equation \eqref{eq4}, but considers the effect of the branching action $a_b$.
The reward function is based on the solving time or the PD gap, so that these two policies can effectively collaborate. The finetuning method for $\pi_c$ is the same as the first learning phase, but in the second phase,  $\pi_b$ also participates in trajectory generation. 
We present the pseudocode of the concurrent joint finetuning phase in Appendix~\ref{app:Pseudocode-2}.

\section{Experiments}
\label{sec5}


In this section, we first introduce the experimental setup, including datasets, implementation details, baselines, and evaluation metrics (Section \ref{51}), followed by the main comparative results (Section \ref{52}). We then evaluate Collab-Solver on challenging instances under an extended time limit (Section \ref{53}) and examine its generalization across instance sizes (Section \ref{56}). Finally, we conduct ablation studies to verify each component (Section \ref{54}) and extend Collab-Solver to collaboration among three modules (Section \ref{collab_three}).

\subsection{Experiment Setup}
\label{51}

\subsubsection{Benchmark Datasets}
We evaluate Collab-Solver on six NP-hard MILP benchmarks: Set Covering \cite{balas1980set}, Maximum Independent Set \cite{bergman2016decision}, Combinatorial Auction \cite{leyton2000towards}, Capacitated Facility Location \cite{cornuejols1991comparison}, Mixed Integer Knapsack \cite{atamturk2003facets}, and Production Planning. As highlighted in the previous work \cite{10607926}, the last dataset, Production Planning, is from a real-world application, and the rest datasets are widely used synthetic MILP benchmarks.
For the synthetic dataset, we generate $10000$ instances for training, $2000$ instances for validation, and $100$ instances for testing.  For the real-world dataset, we split it by $80\%$, $10\%$, and $10\%$ to construct the training, validation, and test sets. The fine-tuning data are a small amount uniformly resampled from the training set.
As we do not aim to investigate the generalization among heterogeneous problems, for each dataset, we train a separate model independently.
Among the six datasets, the Capacitated Facility Location, Mixed Integer Knapsack, and Production Planning datasets contain both integer and continuous variables, and the rest are binary linear programming problems.
Additional details of the datasets and training process are provided in Appendix~\ref{app2}.
For more implementation details, we provide the code for Collab-Solver in the supplementary material.

\subsubsection{Baselines}
We compare Collab-Solver with the learning-based approaches for MILP and a hyper-parameter tuning method: 
\begin{itemize}
    \item SCIP: The backend solver of our approach, where the hyperparameters are kept as default.
    \item SMAC3: As hyperparameters in SCIP are critical to the solving performance, we compare with tuning the cut selection and branching related hyperparameters in SCIP with a Bayesian optimization based automated hyperparameter tuning approach, SMAC3 \cite{lindauer2022smac3}.
    \item HEM: A cut selection policy learning method with a hierarchical policy structure \cite{10607926}, which is similar to the cut agent in our approach. However, to enable communication between the two agents, Collab-Solver augments the hierarchical cut selection policy structure with a GCNN structure to deal with the MILP features. 
    \item GCNN-B: A prevalent imitation-based branching policy learning method with the GCNN model \cite{gasse2019exact}.
    \item RL-B: A reinforcement learning based branching method \cite{scavuzzo2022learning}, which formulates branching as a tree MDP.
\end{itemize}

\subsubsection{Implementation Details}
 In the experiments, we use SCIP 8.0 as the backend solver, a modern open-source solver widely used in the combinatorial optimization research domain \cite{nair2020solving,turner2022adaptive,huang2022learning}. 
 We have not used commercial solvers as backends, such as CPLEX and Gurobi, since they do not provide open APIs for custom cut selection and branching policies. Besides, directly comparing Collab-Solver with these commercial solvers is unfair, as our backend solver (SCIP) is less efficient than them.
The SCIP parameters are kept as default for all baselines except SMAC3, and the SCIP versions for all the baselines are the same as in Collab-Solver. 
Since our method involves both the cut selection and branching modules, we employ SMAC3 to tune the corresponding parameters, including the activation frequency of different cutting plane methods and prioritization of different branching rules, as a baseline. To summarize, the empirical comparisons are as fair and reproducible as possible. 
In addition, all the advanced features of SCIP, such as presolve and primal heuristics, are enabled to guarantee that our setup is consistent with the practical settings. 
The solving time limit is set to $300$ seconds.
The neural networks are optimized with the ADAM optimizer \cite{kingma2014adam} using the PyTorch library \cite{paszke2019pytorch}. 
For the two-timescale update rule, $\omega_c$ is set as $4$ and $\omega_b$ is set as $1$.

\subsubsection{Evaluation Metrics}
 We adopt two well-established evaluation metrics in the MILP domain, i.e., the average solving time (Time) and the average primal-dual gap integral (PD integral). The lower values of these two metrics indicate better performance. The PD integral metric is defined by the area between the curves of the solver's global primal bound and global dual bound, which is calculated as follows,
\begin{equation}
\int_{t=0}^{T}(c^Tx_t^* -z_t^*)dt,
\end{equation}
where $c$ is the object coefficient vector in Equation \eqref{eq1}, $x_t^*$ is the best feasible solution found at time $t$, and $z_t^*$ is the best dual bound at time $t$.
For the easy benchmark datasets with a relatively short solving time, the Time metric is more important, and for the challenging datasets where most instances cannot be solved within the time limit $T$, the PD integral metric matters more, as it is related to the optimality of the found solution.
Note that the Time metric does not consider the training time of the models, and the metrics are evaluated on the test datasets. 

\begin{table*}[htbp]
\centering
\begin{tabular}{c|cccccc}
\toprule\toprule
&SCIP & SMAC3 & HEM & GCNN-B & RL-B & Ours  \\
\midrule
PD gap & $16.57 \pm 4.71$ & $12.91 \pm 3.53$ &  $14.58 \pm 4.26$& $15.47 \pm 4.09$ & $16.10 \pm 9.78$ & $\bf{10.32 \pm 2.83}$  \\
\bottomrule
\end{tabular}
\caption{PD gaps on the IP dataset (time limit=1000s).}
\label{tab2}
\end{table*}

\begin{table*}[t!]
    \centering
    \begin{tabular}{ccccccc}
        \toprule\toprule
       & \multicolumn{2}{c}{Max Independent Set} & \multicolumn{2}{c}{Combinatorial Auction} &  \multicolumn{2}{c}{Capacitated Facility Location}\\
      &\multicolumn{2}{c}{(n = 400, m = 1953)} & \multicolumn{2}{c}{(n = 1000, m = 385)} & \multicolumn{2}{c}{(n = 20100, m = 20303)}\\
        \midrule
        Method &  Time(s) $\downarrow$ & PD Integral $\downarrow$  & Time(s) $\downarrow$ & PD Integral $\downarrow$ & Time(s) $\downarrow$ & PD Integral $\downarrow$ \\
        \cmidrule(lr){2-3} \cmidrule(lr){4-5} \cmidrule(lr){6-7}
        SCIP       & 
        3.70 $\pm$ 0.17           & 28.59 $\pm$ 1.22         &
         21.98 $\pm$ 0.18          & 107.59 $\pm$ 1.27  & 215.46 $\pm$ 3.40         & 1070.63 $\pm$ 1026  \\
        HEM          & 
        1.15 $\pm$ 0.09         & 12.35 $\pm$ 0.84       & 
         18.44 $\pm$ 0.80        & 91.20 $\pm$ 3.44  & 209.51 $\pm$ 6.26         & 1013.65 $\pm$ 30.79  \\
        GCNN-B    & 
        5.16 $\pm$ 0.26           & 26.65 $\pm$ 1.54       & 
         15.18 $\pm$ 0.16          & 82.23 $\pm$ 0.48    & 188.39 $\pm$ 3.49         & 1000.07 $\pm$ 50.17   \\
        Ours        & 
         $\textbf{0.71 $\pm$ 0.02}$           & $\textbf{9.37 $\pm$ 0.22}$        & 
        $\textbf{12.23 $\pm$ 0.18}$          & $\textbf{68.30 $\pm$ 0.46}$   & $\textbf{183.48 $\pm$ 4.76}$         & $\textbf{952.15 $\pm$ 49.56}$ \\
        \bottomrule
    \end{tabular}
    \caption{Generalization experiment results. The models are those trained in Section \ref{52} and are evaluated in new test sets.}
    \label{tab4}
\end{table*}

\subsection{Comparative Results}
\label{52}

As shown in Table \ref{tab1}, Collab-Solver outperforms these learning-based baselines and the hyperparameter tuning method on the six benchmark datasets. 
$n$ and $m$ denote the average number of variables and constraints of the MILPs in the corresponding datasets, which define the problem scales.
The Time metric indicates that the datasets in the second row are more challenging to solve than those in the first row. On the relatively easy datasets in the first row, Collab-Solver has accomplished almost $60\%$ improvement over SCIP with regards to the Time metric. 
Our improvement over the baselines is more prominent in the second row, which indicates that for the challenging instances, the collaboration between various modules in the MILP solver is more critical.
Note that for $3$ out of $10$ test instances in the real-world dataset (Production Planning), our method has reached the time limit (300s), so for this dataset, the PD integral metric is more important than the Time metric. 
Regarding the PD integral metric, Collab-Solver substantially outperforms the baselines.
The learning-based methods, HEM and GCNN-B, perform better than the default SCIP solver on most datasets, which demonstrates the merits of machine learning in the MILP solving process. 
However, the performance of HEM and GCNN-B is inconsistent across these datasets, which implies that the solving of certain datasets relies more on the cut selection policy, and the performance of other datasets is more dependent on the branching policy. 
The proposed approach, Collab-Solver, can take advantage of both aspects. Through collaborative policy learning of cutting and branching, Collab-Solver obtains impressive performance, especially on the challenging datasets in the second row.
The hyperparameter tuning method, SMAC3, performs better than SCIP with default hyperparameters on most datasets. However, on the Set Covering dataset, the Bayesian optimization method SMAC3 has not found better hyperparameters than the default setting, as the default hyperparameters are tuned with human designer knowledge. 
The reinforcement learning based branching method (RL-B) has generally not achieved better performance than the default SCIP, perhaps due to a lack of knowledge from expert datasets. 
Furthermore, the standard deviations of the RL-B results are relatively large, since RL-based approaches easily suffer from the unstable problem. In contrast, Collab-Solver introduces the two-timescale update rule to stabilize the joint learning.

\subsection{Long-Term Performance}
\label{53}
To evaluate the longer-term performance of our approach, we conduct experiments on the Item Placement (IP) dataset from the NeurIPS ML4CO competition \cite{gasse2022machine}. The IP dataset is extremely difficult, where SCIP cannot solve most instances within a time limit of $1000$ seconds. Therefore, we compare the PD gaps with a longer time limit (1000s) in Table \ref{tab2}.
In the long-term evaluation, our method still shows clear improvement over the baselines. On the challenging IP dataset, the final PD gap of GCNN-branching is slightly larger than that of HEM (the cut-selection method). This phenomenon indicates that for the extremely challenging instances, the cuts play a vital role in decreasing the PD gaps.
The hyperparameter-tuning method outperforms SCIP with default parameters, and the solving performance of RL-B is almost the same as that of SCIP.
This phenomenon is generally consistent with the results in Table \ref{tab1}.
It is noteworthy that conducting experiments on these extremely hard datasets is quite time-consuming, so we only conduct the long-term experiments on one dataset. 



\subsection{Size Generalization}
\label{56}

A competitive advantage of learning-based MILP methods over manually designed heuristics is their ability to generalize. In this subsection, we evaluate the generalization ability of Collab-Solver on three synthetic benchmark datasets. Specifically, we change the variable number $n$ and the constraint numbers $m$ to construct a new test dataset, which results in a different difficulty level from the training dataset. Comparing the $n$ and $m$ in Table \ref{tab4} with those in Table \ref{tab1}, it can be found that the test sets are dramatically different from the training sets. 
Although there are significant differences between the training and test sets, benefiting from neural network capabilities and the close collaboration between cut selection and branching policies, Collab-Solver has achieved the strongest generalization, as shown in Table \ref{tab4}.
Compared with SCIP, Collab-Solver has achieved nearly $50\%$ improvement in solving time.
The learning-based cut-selection methods, HEM and GCNN-B, have generally demonstrated better generalization than the backbone solver SCIP. 

\begin{table*}[htbp]
    \centering
    \begin{tabular}{ccccccc}
        \toprule\toprule
      &  \multicolumn{2}{c}{Set Covering}  & \multicolumn{2}{c}{Max Independent Set} & \multicolumn{2}{c}{Combinatorial Auction} \\
      &  \multicolumn{2}{c}{(n = 1000, m = 500)}  & \multicolumn{2}{c}{(n = 500, m = 1953)} & \multicolumn{2}{c}{ (n = 500, m = 192)} \\
        \midrule
        Method & Time(s) $\downarrow$ & PD Integral $\downarrow$ & Time(s) $\downarrow$ & PD Integral $\downarrow$  & Time(s) $\downarrow$ & PD Integral $\downarrow$ \\
        \cmidrule(lr){2-3} \cmidrule(lr){4-5} \cmidrule(lr){6-7}
        w/o F\&Comm     & 2.58 $\pm$ 0.03         & 50.52 $\pm$ 0.76      & 
         4.58 $\pm$ 0.95         & 23.01 $\pm$ 6.86      & 
         0.98 $\pm$ 0.02         & 12.39 $\pm$ 0.29  \\
        w/o F  & 2.57 $\pm$ 0.04         & 50.31 $\pm$ 0.93   & 
         2.70 $\pm$ 0.48         & 17.42 $\pm$ 3.25   & 
         0.83 $\pm$ 0.01         & 11.61 $\pm$ 0.12     \\
        Ours   & $\textbf{2.45 $\pm$ 0.05}$          & $\textbf{49.72 $\pm$ 0.87}$ & 
         $\textbf{1.41 $\pm$ 0.49}$          & $\textbf{13.19 $\pm$ 0.34}$ & 
         $\textbf{0.78 $\pm$ 0.03}$         & $\textbf{10.70 $\pm$ 0.46}$  \\
        \bottomrule
        \\
        \toprule\toprule
       & \multicolumn{2}{c}{Capacitated Facility Location}  & \multicolumn{2}{c}{MIK} & \multicolumn{2}{c}{Production Planning} \\
        & \multicolumn{2}{c}{ (n = 10100, m = 10203)}  & \multicolumn{2}{c}{(n = 413, m = 346)} & \multicolumn{2}{c}{(n = 3582.25, m = 5040.42)} \\
        \midrule
        Method & Time(s) $\downarrow$ & PD Integral $\downarrow$ & Time(s) $\downarrow$ & PD Integral $\downarrow$  & Time(s) $\downarrow$ & PD Integral $\downarrow$ \\
        \cmidrule(lr){2-3} \cmidrule(lr){4-5} \cmidrule(lr){6-7}
        w/o F\&Comm     & 69.78 $\pm$ 6.71         & 301.34 $\pm$ 37.43      & 
        63.13 $\pm$ 3.21         & 178.17 $\pm$ 13.25      & 
        151.43 $\pm$ 6.47        & 7639.75 $\pm$ 108.66 \\
        w/o F  & 67.26 $\pm$ 4.29        & 298.67 $\pm$ 29.49   & 
         57.92 $\pm$ 2.07         & 170.91 $\pm$ 14.46  & 
         141.24 $\pm$ 0.53        & 7163.26 $\pm$ 32.10    \\
        Ours   & $\textbf{57.10 $\pm$ 5.87}$          & $\textbf{282.99 $\pm$ 5.54}$  & 
        $\textbf{46.17 $\pm$ 7.75}$          & $\textbf{117.78 $\pm$ 13.52}$  & 
         $\textbf{138.85 $\pm$ 0.80}$         & $\textbf{6812.83 $\pm$ 78.78}$  \\
        \bottomrule
    \end{tabular}
    \caption{Ablation study results. w/o F denotes removing the fine-tuning phase, and w/o F\&Comm represents removing both finetuning and the data communication in the pretraining phase.}
    \label{tab3}
\end{table*}

\subsection{Ablation Studies}
\label{54}

As shown in Table \ref{tab3}, we compare Collab-Solver with removing the finetuning phase (w/o F), and removing both the finetuning phase and the data communication in the first learning phase (w/o F\&Comm). Note that the w/o F\&Comm experiment can be regarded as an improved version of the previous multi-agent-based MILP method \cite{jing2024multi}.
The experiment results in Table \ref{tab3} demonstrate the merits of both the stabilized concurrent finetuning phase and the data communication in the pretraining phase. 
Comparing the w/o F\&Comm row with the w/o F row, we find that the data communication in the first learning phase contributes much to the collaboration between cutting planes and branching, leading to a substantial improvement. The comparison between the {w/o F} row and the {Ours} row implies that the stabilized finetuning can further promote the collaboration of these two policies, and achieve additional improvement over solving time and PD integral.




\subsection{Extension to Hierarchical Three-Module Collaboration}
\label{collab_three}

We further examine whether Collab-Solver extends beyond two agents by introducing Predict-and-Search (PnS) \cite{huang2024contrastive} as an upstream primal-heuristic agent. Its variable-fixing decisions create a natural three-level dependency: PnS restricts the search space, cut selection tightens the resulting LP relaxation, and branching explores the search tree. We evaluate this extension on large-scale Unit Commitment, for which PnS is particularly suitable.


\textbf{Predict-and-Search.}
PnS is a learning-based primal heuristic that learns variable-fixing scores from historical solutions and fixes selected binary variables before branch-and-cut. For Unit Commitment, it fixes $7.5\%$ of binary variables to 1 and $7.5\%$ to 0 by adding variables and constraints, increasing the formulation size from $(n,m)=(100471,140605)$ to $(107554,154753)$. Dataset details are provided in Appendix~\ref{app2}.


\textbf{Collaboration Setup.}
Since PnS is conducted before cut selection and branching, we formulate it as the super-leader in the hierarchical Stackelberg game. The PnS policy is first pretrained through contrastive learning and then optimized to collaborate with the cut-selection policy via the Collab-Solver scheme. The branching policy is subsequently incorporated to form the complete three-module collaboration framework.

We set the time limit to $3600$ seconds for both Gurobi and SCIP. Since most instances cannot be solved within this limit, we use the best feasible solution $z^{\text{Gurobi}}$ found by Gurobi as the lower bound. During SCIP solving, an instance is considered solved and the solver is terminated once the objective value reaches $101\%$ of this lower bound. The Solved Ratio is defined as the proportion of instances satisfying this criterion within $3600$ seconds. We report solving time, PD integral, Solved Ratio, and the normalized PD gap:
\begin{equation}
\mathrm{Norm\ PD\ Gap} =
\frac{z^{\text{SCIP}} - z^{\text{Gurobi}}}{z^{\text{SCIP}}}.
\label{eq:pd_gap}
\end{equation}


\textbf{Results.}
As shown in Table~\ref{tab10}, Collab-Solver achieves the best performance across all metrics, reducing solving time by approximately $26\%$ and increasing the Solved Ratio from $57\%$ to $71\%$ relative to SCIP. Its lower PD integral and normalized gap indicate faster convergence and better final solution quality. Collab-Solver solves $5$ of the $7$ test instances, compared with $4$ for all other methods, while the single-module learning-based methods also slightly outperform SCIP. These results provide a proof of concept for hierarchical three-module collaboration.

\begin{table}[t!]
    \centering
    \resizebox{\linewidth}{!}{%
    \begin{tabular}{ccccc}
        \toprule\toprule
        &\multicolumn{4}{c}{Unit Commitment}   \\
        \midrule
        Method & Time(s) $\downarrow$ & PD Integral $\downarrow$  & Norm PD Gap $\downarrow$ & Solved Ratio $\uparrow$ \\
         \cmidrule(lr){2-5}
        SCIP     & 2226 $\pm$ 24         & 124086 $\pm$ 760       & 
         1.48\%  $\pm$ 0.00\%    &    57\%  $\pm$ 0.00\%  \\
        HEM     & 2199 $\pm$ 26         & 124801 $\pm$ 1551     & 
        1.41\% $\pm$ 0.00\%      &   57\% $\pm$ 0.00\%           \\ 
        GCNN-B  & 2195 $\pm$ 27        & 121190 $\pm$ 1877     & 
        1.48\% $\pm$ 0.00\%    &     57\%  $\pm$ 0.00\%          \\
         PnS     &  2037 $\pm$ 22         & 103687 $\pm$ 1232      & 
         1.19\% $\pm$ 0.00\%   &      57\%   $\pm$ 0.00\%   \\
        Ours    & $\textbf{1632 $\pm$ 17}$          & $\textbf{99391$\pm$ 1741}$      & 
        $\textbf{1.15\% $\pm$ 0.00\%}$       &   \textbf{71\%   $\pm$ 0.00\%}     \\
        \bottomrule
        
    \end{tabular}
    }
    \caption{ The three-module collaboration results. 
 The best performance is marked in bold. Each experiment has been run with $5$ random seeds, and the mean and std are reported.
    }
    \vspace{-20pt}
    \label{tab10}
\end{table}



\section{Conclusion}

We proposed Collab-Solver, a collaborative multi-agent framework for learning-enhanced MILP solving that jointly optimizes cut selection and branching through data-communicated pretraining and concurrent online fine-tuning. Our results show that independently trained solver modules may collaborate suboptimally, whereas explicitly modeling their leader--follower interaction enables more effective coordination. We further formulate joint optimization as a non-stationary learning problem and employ a two-timescale update rule to stabilize concurrent training. Extensive experiments across diverse MILP datasets demonstrate consistent improvements over the evaluated learning-based policies and hyperparameter-tuning baseline. Since Collab-Solver currently trains a separate model for each problem class, future work will explore multi-task learning to improve cross-class generalization.


\begin{acks}
The authors thank the anonymous reviewers for their valuable comments and helpful suggestions. 
This work was supported by the National Natural Science Foundation of China (62306088) and the Natural Science Foundation of Heilongjiang Province (YQ2024F007).
\end{acks}






\bibliographystyle{ACM-Reference-Format}
\bibliography{sigconf}

\clearpage
\appendix

\section{Algorithm Details}
\label{app1}
\subsection{Details of the Branching Policy}
\label{s_b}
As shown in Equation \eqref{imi_loss}, the branching policy $\pi_b$ is optimized with the cross-entropy loss with a minibatch size of $32$ and a learning rate of $0.001$. We adopt the branching feature representation $o_b=(C,E,V)$ as that in the GCNN-B method \cite{gasse2019exact}.
The action $a_c$ is implicitly included in the constraint features $C$.
The details of the bipartite feature representation $o_b$ are listed in Table \ref{tab_gcnn}.
\begin{table*}[htbp]
    \footnotesize
	\centering
	\begin{tabular}{ccc}
        \toprule
		\textbf{Notation} & \textbf{Feature} & \textbf{Description}  \\
        \midrule
		\multirow{5}{*}{\textbf{C}} & obj\_cos\_sim & Cosine similarity with objective. \\
		\cline{2-3}
		& bias & Bias value, normalized with constraint coefficients \\
		\cline{2-3}
		& is\_tight & Tightness indicator in LP solution. \\
		\cline{2-3}
		& dualsol\_val & Dual solution value, normalized. \\
		\cline{2-3}
		& age & LP age, normalized with the total number of LPs. \\
		\hline
		\multirow{1}{*}{\textbf{E}} & coef & Constraint coefficient, normalized per constraint. \\
		\hline
		\multirow{13}{*}{\textbf{V}} & type & Type (binary, integer, impl. integer, continuous) as a one-hot encoding.  \\
		\cline{2-3}
		& coef & Objective coefficient, normalized. \\
		\cline{2-3}
		& has\_lb & Lower bound indicator. \\
		\cline{2-3}
		& has\_ub & Upper bound indicator. \\
		\cline{2-3}
		& sol\_is\_at\_lb & Solution value equals lower bound. \\
		\cline{2-3}
		& sol\_is\_at\_ub & Solution value equals upper bound. \\
		\cline{2-3}
		& sol\_frac & Solution value fractionality. \\
		\cline{2-3}
		& basis\_status & Simplex basis status (lower, basic, upper, zero) as a one-hot encoding. \\
		\cline{2-3}
		& reduced\_cost & Reduced cost, normalized. \\
		\cline{2-3}
		& age & LP age, normalized \\
		\cline{2-3}
		& sol\_val & Solution value. \\
		\cline{2-3}
		& inc\_val & Value in incumbent. \\
		\cline{2-3}
		& avg\_inc\_val & Average value in incumbents. \\
        \bottomrule
	\end{tabular}
    \caption{Description of the constraint, edge, and variable features in our bipartite state representation $o_b=(C,E,V)$. This representation corresponds to the MILP features, which describe most of the current node in the B\&B tree.}
	\label{tab_gcnn}
\end{table*}

\subsection{Details of the Cut Selection Policy}
The cut selection agent utilizes the reinforcement learning method to train a hierarchical policy.
The input of the cut selection policy $\pi_c$ includes two parts: the MILP features $o_b$ and the candidate cut features $o_c$. 
The MILP features are the same as those in the previous subsection (Sec \ref{s_b}).
Inspired by the HEM method, we encode each cutting plane as a $13$-dimensional feature vector for representation, which composes the cut features $o_c$.
The specific meaning of each element of the cut feature $o_c$ is detailed in Table \ref{app_tab4}. 
The higher-level network in the hierarchical policy structure outputs a ratio \(k\), representing the proportion of selected cutting planes. The lower-level policy is encoded with a pointer network, and outputs the selected cuts $a_c$.

\begin{table*}[ht]
\footnotesize
    \centering
    \begin{tabular}{ccc}
        \toprule
        \multicolumn{1}{c}{Feature}  & \multicolumn{1}{c}{Description} & \multicolumn{1}{c}{Value} \\
        \midrule
        cut coefficients & the mean, max, min, std of cut coefficients  & 4 \\
        objective coefficients & the mean, max, min, std of the objective coefficients & 4 \\
        parallelism & the parallelism between the objective and the cut  & 1 \\
        efficacy & the Euclidean distance of the cut hyperplane to the current LP solution & 1 \\
        support & the proportion of non-zero coefficients of the cut & 1 \\
        integral support & the proportion of non-zero coefficients with respect to integer variables of the cut & 1 \\
        normalized violation & the violation of the cut to the current LP solution  & 1 \\
        \bottomrule
    \end{tabular}
    \caption{The designed cut features of a candidate cut.}
    \label{app_tab4}
\end{table*}

\subsection{Reward Design}

The reward function for $\pi_c$ and $\pi_b$ is the negative of the time interval from the start of the solving process to the current time in all the datasets except Production Planning, IP, and Unit Commitment. The reward function is shown as follows.
\begin{equation}
 r_t = -(t-t_0).
 \label{rb}
\end{equation}
Note that the non-zero rewards are only given when the solving process ends.
Regarding the challenging datasets (Production Planning, IP, and Unit Commitment), which have hit the time limit, we use the negative of the current PD gap as the reward. Similarly, for the intermediate timestep, the rewards are zeros.
For more implementation details, we provide the code for Collab-Solver in the supplementary material.

\section{Experiment Details}
\label{app2}
\subsection{Datasets}
We present the details about the benchmark datasets below. Readers can refer to \url{https://atamturk. ieor.berkeley.edu/data/mixed.integer.knapsack/} for the Mixed Integer Knapsack dataset.

\textbf{Set Covering}

Given the elements $1, 2,..., m$, and a collection $\mathcal{S}$ of $n$ sets whose union equals the set of all elements, the set cover problem can be formulated as follows:

\begin{equation}
    \begin{aligned}
    \min &\sum_{s\in \mathcal{S}} x_s\\
    \text{s.t. } &\sum_{s:e\in s} x_s \geq 1,\ e=1,..., m\\
    &x_s\in \{0, 1\},\ \forall s\in \mathcal{S}
    \end{aligned}
\end{equation}

\textbf{Max Independent Set}

Given a graph $G$, the Max Independent Set problem consists of finding a subset of nodes of maximum cardinality so that no two nodes are connected. We use the clique formulation from \cite{bergman2016decision}. Given a collection $\mathcal{C} \subset 2^V$ of cliques whose union covers all the edges of the graph $G$, the clique cover formulation is 
\begin{equation}
\begin{aligned}
    \max &\sum_{v\in V}x_v \\
    s.t. &\sum_{v \in C} x_v \leq 1, \forall C \in \mathcal{C}\\
    &x_v \in \{0, 1\}, \forall v \in V
\end{aligned}
\end{equation}

\textbf{Combinatorial Auction}

For $m$ items, we are given $n$ bids $\{\mathcal{B}_j\}^{n}_{j=1}$. Each bid $\mathcal{B}_j$ is a subset of the items with an associated bidding price $p_j$. The associated combinatorial auction problem is as follows:
\begin{equation}
    \begin{aligned}
        \max &\sum_{j=1}^n p_jx_j \\
        s.t. &\sum_{j:i\in \mathcal{B}_j} x_j \leq 1, i=1,..., m\\
        &x_j \in \{0, 1\}, j=1, ..., n
    \end{aligned}
\end{equation}
where $x_j$ denotes the action of choosing bid $\mathcal{B}_j$.

\textbf{Capacitated Facility Location}

Given a number $n$ of clients with demands $\{d_j\}_{j=1}^n$, and a number of $m$ of facilities with fixed operating costs $\{f_i\}_{i=1}^m$ and capacities $\{s_i\}_{i=1}^m$, let $c_{ij}/d_j$ be the unit transportation cost between facility $i$ and client $j$, and let $p_{ij}/d_j$ be the unit profit for facility $i$ supplying client $j$. The MILP problem is as follows,

\begin{equation}
    \begin{aligned}
        min& \sum_{i=1}^m\sum_{j=1}^{n}c_{ij}x_{ij} + \sum_{i=1}^{m}f_iy_i \\
        s.t. &\sum_{j=1}^n d_jx_{ij}\leq s_iy_i, \  i=1,..., m\\
        &\sum_{i=1}^{m}x_{ij}\geq 1, \  j=1, ..., n\\
        &x_{ij} \geq 0 \ \forall i, j\\
        &y_i \in \{0, 1\} \ \forall i
    \end{aligned}
\end{equation}
where each variable $x_{ij}$ represents the decision of facility $i$ supplying client $j$'s demand, and each variable $y_i$ denotes the decision of opening facility $i$ for operation.

\textbf{Production Planning}

The production planning problem aims to find the optimal production plan for thousands of factories according to the daily demand for orders. The constraints include the production capacity for each production line in each factory, transportation limit, the order rate, etc. The optimization objective is to minimize the production cost and time simultaneously. 
This data set is adopted from the HEM paper \cite{wanglearning}.
The average size of the production planning problems is approximately equal to $3500 \times 5000 = 1.75 \times 10^8$, which are large-scale real-world problems. 

\textbf{Unit Commitment}

Consider a power system with $N_B$ buses, $N_G$ generators and $N_L$ transmission lines. Set $B=\{1,2,3,..., N_B\}$, $G=\{1,2,3,..., N_G\}$,  $L=\{1,2,3,..., N_L\}$, and  $D=\{1,2,3,..., N_D\}$ represent the sets of buses, generators, lines, and nodal demands respectively. For each generator $g\in G$,
let a binary variable $I_g \in \{0, 1\}$ represent the on/off status of the generator $g$ and a continuous variable $P_g$ indicate its generated power. A linear-cost typical unit commitment problem is formulated as follows.

\begin{equation}
    \begin{aligned}
        min& \sum_{g\in G} b_{g}P_{g} + a_gI_g \\
        s.t. &\sum_{g\in G_b}  P_g=d_b+ \sum_{l\in L_b} P_l   &\forall b \in B\\
        & -\overline{P_l}\leq P_l \leq \overline{P_l}  &\forall l\in L \\
        & \underline{P_g}I_g \leq P_g \leq \overline{P_g}I_g &\forall g\in G \\
        &\sum_{g\in G}R_g\geq R  \\
       & P_g + R_g \leq \overline{P_g}I_g & \forall g \in G\\
        &I_g \in \{0, 1\} & \forall g \in G
    \end{aligned}
\end{equation}

Here $d_b$ is the forecasted net power demand at bus $b$, $P_l$ denotes the active power flow on line $l$, and $\overline{P_l}$ is the maximum power capacity. $\overline{P_g}$, $\underline{P_g}$ and $R_g$ are the maximum and minimum power and the spinning reserve of generator $g$ respectively. $R$ is the system requirement.

\subsection{Hyperparameters}
We present the hyperparameter settings throughout the experiments in three distinct tables. Each table lists the parameters in different learning phases of the proposed method to enhance reproducibility.

\textbf{Hyperparameters for Cut Selection Policy Learning.}
 Table \ref{tab6} presents the hyperparameters in the cut selection policy pretraining.
\begin{table}[H]  
\footnotesize
    \centering
    \begin{tabular}{ccc}
        \toprule
        \multicolumn{1}{c}{Symbol}  & \multicolumn{1}{c}{Description} & \multicolumn{1}{c}{Value} \\
        \midrule
        $\mathcal{T}_{limit}$ & Time limit for SCIP  & 300 \\
        $\mathcal{R}_{max}$ & Maximum rounds for cut selection on each node & 1 \\
        $\eta_{l}$ & Learning rate for low-level policy  & $1 \times 10^{-4}$ \\
        $\eta_{h}$ & Learning rate for high-level policy & $5 \times 10^{-4}$ \\
        $\Delta_{lr}$ & Step size for learning rate decay & 5 \\
        $\eta_{lr}$ & Rate of learning rate decay & 0.96 \\
        ${N}_{c}$ & Number of training epochs & 100 \\ 
        ${G}_{e}$ & Maximum gradient norm & 2.0 \\
        ${N}_{batch}$ & Batch size & 16 \\
        $\beta_{critic}$ & Smoothing factor & 0.9 \\
        ${N}_{jobs}$ & Number of parallel jobs for training and testing & 2 \\
        ${d}_{h}$ & Dimension of the hidden layer & 128 \\
        ${N}_{g}$ & Number of glimpses in attention mechanism & 1 \\
        ${B}_{size}$ & Size of beam search & 1 \\
        $\rho$  & Ratio of high-level and low-level updates & 1 \\
        ${F}_{t}$ & Frequency of testing & 3 \\       
        \bottomrule
    \end{tabular}
    \caption{Hyperparameters for $\pi_{c}$ pretraining.}
    \label{tab6}
\end{table}

\begin{table}[htbp]  
\footnotesize
    \centering
    \begin{tabular}{ccc}
        \toprule
      Symbol &  Description & Value \\
        \midrule
       $\mathcal{T}_{limit}$ & Time limit for SCIP  & 300 \\
       $N$ & Expert dataset size & 5000 \\
       $N_b$ & Number of training epochs & 100 \\
       $d_h$ & Dimension for hidden layer & 128 \\
      $f$ &  Activation function & Tanh \\
        $N_{batch}$ & Batch size & 32 \\
    $\eta_b$ &     Learning rate & $1 \times 10^{-3}$\\     
        ${p}_{s}$ & Sample probability & 0.1 \\
        \bottomrule
    \end{tabular}
    \caption{Hyperparameters for $\pi_b$ pretraining.}
    \label{tab7}
\end{table}

\textbf{Hyperparameters for Branching Policy Learning.}
Table \ref{tab7} lists the hyperparameters of $\pi_b$ pretraining.
The hyperparameters for the branching policy $\pi_b$ are fewer, as $\pi_b$ is pretrained with imitation learning.

\textbf{Hyperparameters for Concurrent Joint Finetuning.}
Table \ref{tab8} lists the hyperparameters in the concurrent joint finetuning phase.
For the hyperparameters consistent with those in Table \ref{tab6} and Table \ref{tab7}, we have not relisted them in Table \ref{tab8}.
\begin{table}[H]  
    \footnotesize
    \centering
    \begin{tabular}{ccc}
        \toprule
        \multicolumn{1}{c}{Symbol}  & \multicolumn{1}{c}{Description} & \multicolumn{1}{c}{Value} \\
        \midrule
        ${N}_{f}$ & Number of finetuning epochs & 35 \\
        $\eta$ & Learning rate & $1 \times 10^{-5}$ \\
        $\gamma$ & Discount factor & 0.99 \\
        $\epsilon$ & Exploration probability & 0.05 \\
        $\omega_{c}$ & Update interval of $\pi_c$ & 4 \\
        $\omega_{b}$ & Update interval of $\pi_b$ & 1 \\        
        \bottomrule
    \end{tabular}
    \caption{Hyperparameters for concurrent joint finetuning.}
    \label{tab8}
\end{table}

\section{Pseudocode}
\label{app:Pseudocode}

\subsection{Pseudocode of Data Communicated Pretraining}
\label{app:Pseudocode-1}
\begin{algorithm}[h]
\footnotesize
    \caption{Data Communicated Pretraining }
    \label{alg:data communicated pretraining}
    \begingroup
        \setlength{\parindent}{0pt}
        \setlength{\leftskip}{0pt}
        \setlength{\parskip}{0pt}
        \raggedright
    \textbf{Input}: MILP instances $\mathcal{D}$ for training, 
        number of training epochs  $N_c$, 
        number of training epochs $N_b$,
        expert dataset size $N$, probability of strong branching $p_s$\\
        \textbf{Initialize}: cut selection policy $\pi_c[\theta]$, variable selection policy $\pi_b[\psi]$, cut selection training buffer $\mathcal{D}_c$, expert dataset for branching $\mathcal{D}_e$\\
        \textbf{Output}: Policies $\pi_c[\theta]$ and $\pi_b[\psi]$ \par
    \endgroup
    \begin{algorithmic}[1] 
        \FOR{each training epoch in $N_c$}
            \STATE Sample a MILP from $\mathcal{D}$ and start solving in SCIP
            \WHILE{solving is not complete and not hit the time limit}
                \STATE Use $\pi_c$ with $\epsilon$-greedy for cut selection
            \ENDWHILE
            \STATE Store transitions and rewards for cut selection in $\mathcal{D}_c$
            \STATE Update $\theta$ using policy gradient method with the data in $D_{c}$
        \ENDFOR
        \WHILE{$|\mathcal{D}_{e}| < N$}
            \STATE Sample a MILP from $\mathcal{D}$ and start solving in SCIP
            \WHILE{solving is not complete and not hit the time limit}
            \STATE Use the pretrained $\pi_c$  for cut selection
                \STATE Generate a random number $p$ that satisfies $0 \leq p \leq 1$
                \IF{$p < p_s$} 
                \STATE Branch with \textbf{strong branching} and store the expert data to $\mathcal{D}_e$
                \ELSE
                    \STATE Branch with the default method in SCIP
                \ENDIF
            \ENDWHILE
        \ENDWHILE
        \STATE Update $\psi$ with the data in $D_{e}$ for $N_b$ epochs
        \STATE \textbf{return} $\pi_c[\theta]$ and $\pi_b[\psi]$
    \end{algorithmic}
\end{algorithm}
Lines 1-8 correspond to the pretraining of the cut selection policy $\pi_c$, where cut selection is conducted throughout the whole solving process. The B\&C search tree will not grow exponentially, as the solution can be obtained or a time limit can be hit. 
Lines 9-22 aim to train the branching policy $\pi_b$, which is influenced by the pretrained $\pi_c$ via data communication. 
Note that in Lines 13-18 of Algorithm \ref{alg:data communicated pretraining}, we stochastically call strong branching to diversify the training data for $\pi_b$.

\subsection{Pseudocode of Concurrent Joint Finetuning}
\label{app:Pseudocode-2}
As shown in Lines 3-6, these two policies concurrently involve the solving process of an instance and communicate with each other through the MILP features of bipartite graphs. 
The collected training data are stored in separate buffers $\mathcal{D}_c$ and $\mathcal{D}_b$.
In Lines 9-14, the cutting and branching policies are fine-tuned with the two-timescale update rule to stabilize the training. 
The more frequent updating of the follower policy $\pi_b$ also considers the backward induction in the deviation of the Stackelberg equilibrium, where the follower's best response is obtained before the leader policy optimization. 
\begin{algorithm}[h!]
\footnotesize
    \caption{Concurrent Joint Finetuning }
    \label{alg:concurrent joint finetuning}
    \begingroup
        \setlength{\parindent}{0pt}
        \setlength{\leftskip}{0pt}
        \setlength{\parskip}{0pt}
        \raggedright
    \textbf{Input}: MILP instances $\mathcal{D}$, pretrained cut selection policy $\pi_c[\theta]$, pretrained branching policy $\pi_b[\psi]$, number of finetuning epochs $N_f$ \par
    \textbf{Initialize}: training buffer $\mathcal{D}_{c}$ for $\pi_c[\theta]$, training buffer $\mathcal{D}_{b}$ for $\pi_b[\psi]$ \par
    \noindent \textbf{Output}: Policies $\pi_c[\theta]$ and $\pi_b[\psi]$ \par
    \endgroup
    \begin{algorithmic}[1] 
        \FOR{$n$ in $N_f$}
            \STATE Sample a MILP from $\mathcal{D}$ and start solving
            \WHILE{solving is not complete}
                \STATE Use $\pi_c$ with $\epsilon$-greedy for cut selection
                \STATE Use $\pi_b$ with $\epsilon$-greedy for branching
            \ENDWHILE
            \STATE Store transitions and rewards for cut selection in $\mathcal{D}_c$
            \STATE Store transitions and rewards for branching in $\mathcal{D}_e$
            \IF{$n \mod \omega_b == 0$}
                \STATE Update $\psi$ using policy gradient method based on $\mathcal{D}_{b}$
            \ENDIF
            \IF{$n \mod \omega_c == 0$}
                \STATE Update $\theta$ using policy gradient method based on $\mathcal{D}_{c}$
            \ENDIF
        \ENDFOR
        \STATE \textbf{return} $\pi_{c}[\theta]$ and $\pi_{b}[\psi]$
    \end{algorithmic}
\end{algorithm}

\section{Additional Results}

\begin{figure}[htbp]
    \centering
    \includegraphics[width=0.95\linewidth]{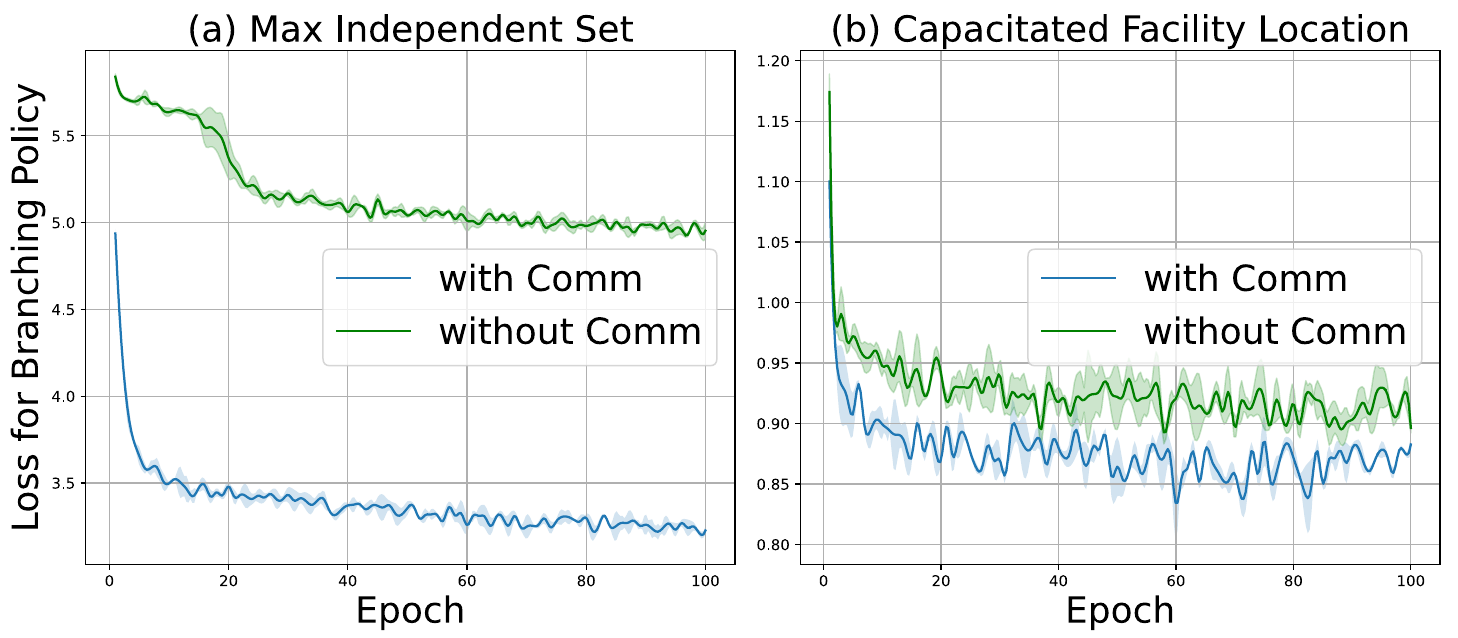}
    \caption{Comparing the branching policy pretraining processes with data communication and without data communication.}
    \label{fig3}
\end{figure}

We provide the comparison of \textit{with data communication} and \textit{without data communication} for the branching policy pretraining processes in Figure \ref{fig3}. As shown by the training loss curves, through data communication, the convergence of imitation-based branching policy learning is faster. Furthermore, the training loss of data communication converges to a smaller value, which indicates a better convergent model for branching. 
Note that the initial training loss of data communication is smaller than that of no data communication in Max Independent Set, which also implies the effectiveness of data communication in expert dataset construction.

\begin{figure}[htbp]
    \centering
    \includegraphics[width=0.85\linewidth]{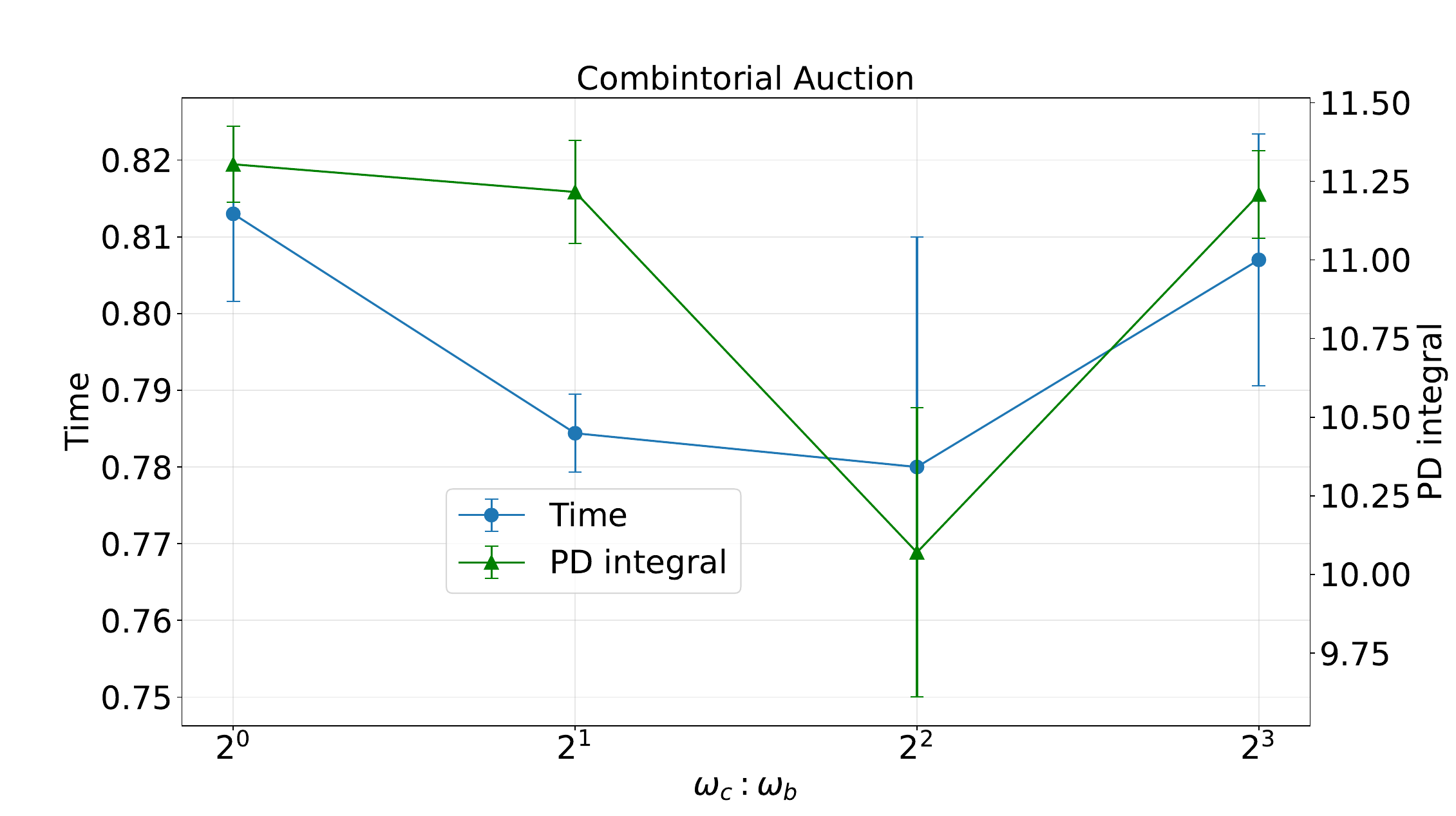}
    \caption{The hyperparameter study results. The mean and standard deviation of solving time and PD integral under different hyperparameter settings are shown above.}
    \label{fig4}
\end{figure}

Beyond the ablation studies, we perform a hyperparameter study on the Combinatorial Auction dataset to evaluate  the hyperparameters of the two-timescale update rule in the finetuning phase, $\omega_c$ and $\omega_b$. Specifically, we vary these two hyperparameters with four different sets of values, in an exponential upward trend with $2$ as the base.
The solving time and PD integral under different hyperparameter settings are shown in Figure \ref{fig4}.
The performance of $\omega_c :\omega_b = 2, 4, 8$ is better than that of $\omega_c : \omega_b = 1$, which validates the effectiveness of the two-timescale update rule in the finetuning phase. $\omega_c : \omega_b=1$ indicates that the leader policy and the follower policy are updated at the same frequency, which leads to the unstable simultaneous policy learning for these two modules. 

\end{document}